\documentclass[letterpaper, 10 pt, fleqn, conference, twoside]{ieeeconf}  % Comment this line out if you need a4paper
\IEEEoverridecommandlockouts                              % This command is only needed if you want to use the \thanks command
\usepackage{capt-of,graphicx,subcaption}
\usepackage{xargs}                      % Use more than one optional parameter in a new commands
\usepackage[pdftex,dvipsnames,table,xcdraw]{xcolor}  % Coloured text etc.
\usepackage{multirow}
\usepackage[colorinlistoftodos,prependcaption]{todonotes}
\usepackage[hidelinks]{hyperref}
\hypersetup{
  colorlinks   = true, %Colours links instead of ugly boxes
  urlcolor     = blue, %Colour for external hyperlinks
  linkcolor    = blue, %Colour of internal links
  citecolor   = red %Colour of citations
}

\usepackage{amsmath,amssymb,bm,mathtools}

\usepackage{makecell}
\usepackage{cite}
\usepackage{multirow}
\usepackage[font={small,it}]{caption}
\usepackage{relsize}
\usepackage{tikz}
\usetikzlibrary{shadings}
\usepackage[margin=1.92cm]{geometry} %% Modify page margins if needed
\usepackage{algorithm}
\usepackage[noend]{algpseudocode}
\usepackage[subpreambles]{standalone} %% include standalone pseudocodes
\newcommand{\subparagraph}{}

\usepackage{array}
\usepackage{booktabs}
\usepackage{epstopdf}

\usepackage[utf8]{inputenc}
\usepackage{pgfplots}
\pgfplotsset{compat=newest}
\usepgfplotslibrary{groupplots}
\usepgfplotslibrary{dateplot}

\makeatletter
\def\BState{\State\hskip-\ALG@thistlm}
\makeatother

\newcommand{\Method}{MuPNet} %% acronym
\newcommand{\MethodFull}{Multi-modal Predictive Coding Network} %% Full-form
\graphicspath{{./sections/figures/}} % Define path to folder containing figures

\title{\LARGE \textbf{\Method: \MethodFull~for Place Recognition by Unsupervised Learning of Joint Visuo-Tactile Latent Representations}}

\author{Oliver Struckmeier, Kshitij Tiwari, Shirin Dora, Martin J. Pearson,\\Sander M. Bohte, Cyriel MA Pennartz and Ville Kyrki
\thanks{This research has received funding from the European Union’s Horizon 2020 Framework Programme for Research and Innovation under the Specific Grant Agreement No. 785907 (Human Brain Project SGA2).}
\thanks{Oliver Struckmeier, Kshitij Tiwari, and Ville Kyrki are with the Department of Electrical Engineering and Automation, Aalto University, Espoo 02150, Finland \tt \footnotesize \{firstname.lastname\}@aalto.fi}
\thanks{Shirin Dora (\texttt{\footnotesize s.dora@uva.nl}) and Cyriel MA Pennartz (\texttt{\footnotesize c.m.a.pennartz@uva.nl}) are with the Cognitive and Systems Neuroscience Group, University of Amsterdam}
\thanks{Sander M Bohte (\texttt{\footnotesize S.M.Bohte@cwi.nl}) is with Centrum Wiskunde \& Informatica, Amsterdam}
\thanks{Martin J. Pearson is with the Bristol Robotics Laboratory, Bristol BS16 1QY, U.K (\tt\footnotesize martin.pearson@brl.ac.uk)}
}

\begin{document}
\maketitle
\begin{abstract}
Extracting and binding salient information from different sensory modalities to determine common features in the environment is a significant challenge in robotics.
Here we present \Method \ (\MethodFull), a biologically plausible network architecture for extracting joint latent features from visuo-tactile sensory data gathered from a biomimetic mobile robot.
In this study we evaluate \Method \ applied to place recognition as a simulated biomimetic robot platform explores visually aliased environments.
The F1 scores demonstrate that its performance over prior hand-crafted sensory feature extraction techniques is equivalent under controlled conditions, with significant improvement when operating in novel environments.
\end{abstract}
%% abstract

%% Introduction
\section{Introduction}
Place recognition is an important ability for autonomous systems that navigate and interact with their environment.
The core requirements for a successful place recognition, such as evaluating the similarity between scenes and comparing them to a set of internal representations, have been extensively researched in recent computer vision and robotics literature.
The recent advances in visual sensors, computer vision and deep learning research have shifted the focus of previous research on place recognition towards using vision as the primary sensory modality~\cite{lowry2015visual}.
However, the main challenge such as recognizing places in changing, cluttered or aliased environments is yet to be fully solved and is an ongoing research~\cite{chen2014convolutional,kendall2015posenet}.

\begin{figure}[htp]
\centering
\includegraphics[width=\columnwidth]{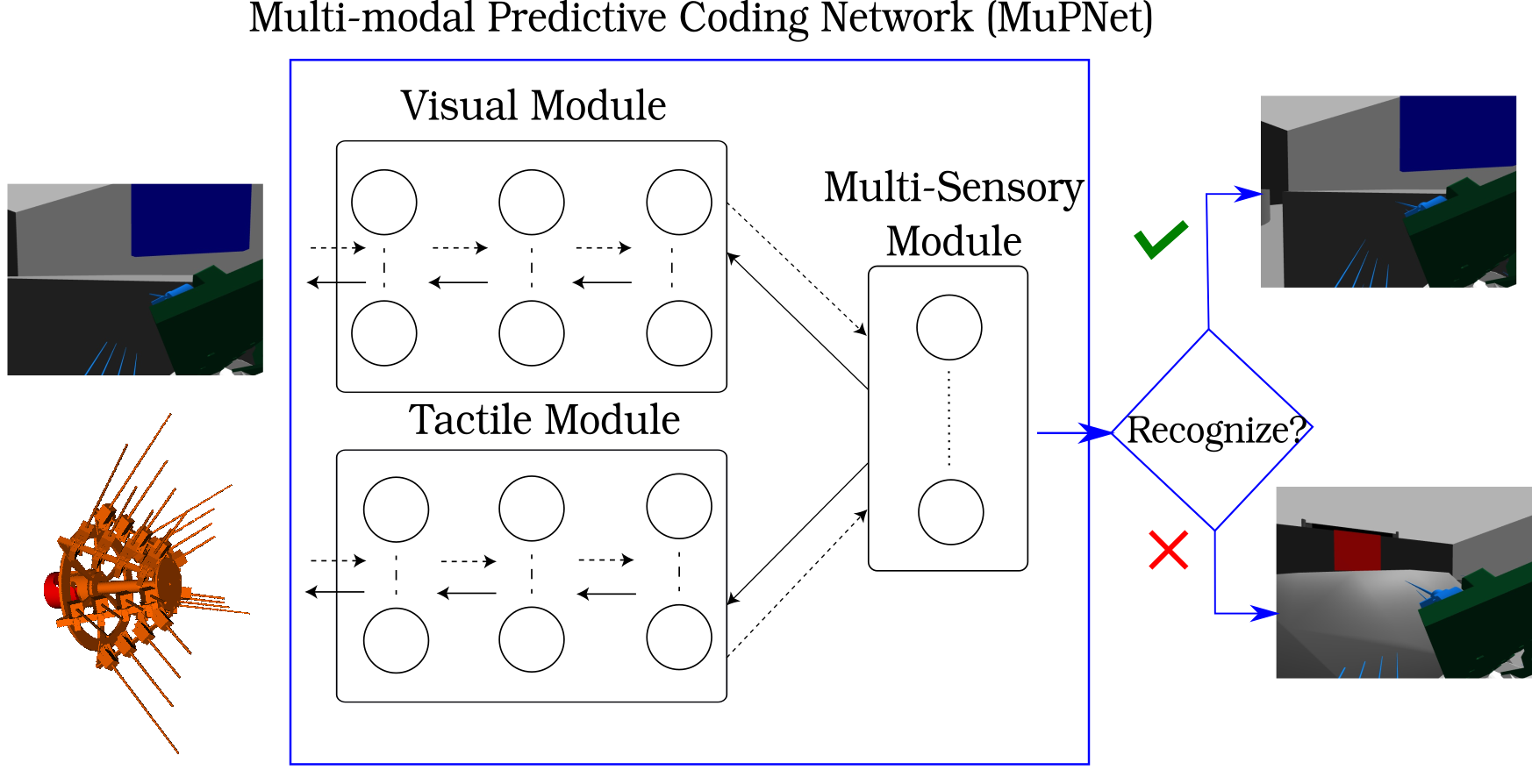}
\caption{High level overview of \Method~architecture. Visuo-tactile sensory data for a given scene is presented to the trained network and a latent representation of the multi-sensory stimuli is inferred. Note that the errors (dashed lines) are propagated forward and the predictions (solid lines) backwards. The resulting representation is used for place recognition.}
\label{fig:high_level}
\end{figure}

Previous works such as \cite{milford2013brain} have shown that fusing multiple sensory modalities which complement each other improves place recognition and simultaneous localization and mapping (SLAM) performance, especially in cluttered and aliased environments.
In such environments, tactile sensors can interact with the surrounding in close range and help discern ambiguous visual landmarks and prevent wrong place recognitions~\cite{struckmeier2019vita}.
Recently, new tactile sensors have pushed the precision of spatio-temporal acuity to the level of a human finger tip \cite{yuan2017gelsight,Oddo2011roughness,lepora2015super} and it has been shown that tactile sensing can be used for close range object recognition \cite{salman2018whisker,Bauza2019TactileMA}.
For obtaining information about the geometry of objects, bio-mimetic rat whiskers are capable to provide a robust measure of surface proximity, information from such arrays can be used to determine surface form, texture, compliance and friction \cite{kim2007biomimetic, pearson2007whiskerbot,solomon2006biomechanics}.
However, extraction of features from these sensors is typically performed using hand-crafted features, which has been found inferior in the context of visual processing, where convolutional neural network architectures are used to learn the features.
Moreover, the existing features represent typically only visual or tactile signatures, and the typical approach to combine them by weighting does not address their correlations.

We propose a new biologically plausible feature extraction method called  \Method \ (\MethodFull) that implicitly fuses visual and tactile information into a single feature. 
The method uses neurobiologically plausible predictive coding illustrated in Fig.~\ref{fig:high_level} to infer latent visuo-tactile representations of the sensory input. 
Using three settings, we empirically study the robustness of place recognition with the developed features compared to hand-crafted sensory pre-processing techniques adopted in prior works~\cite{struckmeier2019vita}.
Although this study concerns place recognition using vision and touch, we contend that \Method \ is applicable to learning the joint latent representations from any co-incident multi-sensory input.

The main contributions of the work are: (i) extension of predictive coding to multi-modal sensory information; (ii) the method for biologically plausible visuo-tactile feature extraction; and (iii) demonstration of improved robustness of place recognition compared to prior works when faced with contextual changes.

Previous research such as \cite{lee2019making} have inferred representations of multiple sensory modalities using a hierarchical autoencoder with individual encoder/decoder blocks for each sensory modality.
One of the main differences between predictive coding and existing machine learning models like an autoencoder is the direction in which information and errors propagate.
An autoencoder consists of an encoder and a decoder which together form a feedforward network which is trained end-to-end using error-backpropagation.
However, error backpropagation is biologically implausible~\cite{lillicrap2016random} and predictive coding is a biologically plausible alternative.
During inference autoencoders, propagate information sequentially towards the output layer in the network whereas in predictive coding all layers in the network parallelly transmit information only towards the input layer (without any further propagation across layers). For learning, autoencoders require a backward-pass through the network from output to input layer whereas in predictive coding each layer parallelly transmits prediction errors towards the multi-sensory module as shown in Fig. \ref{fig:high_level}.
Furthermore, in autoencoders, neuronal activity in intermediate layers is derived from the feedforward propagation of the input. In predictive coding, the neuronal activity in each layer is initialized randomly and then adapted such that it best represents the features of a given multi-modal input. Thus, the predictive coding architecture can infer representations without an explicit encoding block.

%% Prelims
\section{Multi-modal feature extraction}
We begin this section by presenting an existing hand-crafted baseline that has been proposed for bio-inspired SLAM. We then continue by presenting the predictive coding based features.

\subsection{Hand-crafted baseline}
\begin{figure}[tbp]
\centering
\resizebox{.4\textwidth}{!}{\begin{tikzpicture}[
				round boxes/.style={draw, rectangle,%
                thick,minimum height=1cm, rounded corners,
                minimum width=1cm, black, text=black,
                text width=15mm, anchor=center, align=center},
                boxes/.style={draw, rectangle,%
                thick,minimum height=1cm,
                minimum width=1cm, black, text=black,
                text width=15mm, anchor=center, align=center}, 
    			% Define styles for some special nodes
   				right iso/.style={isosceles triangle,scale=1.0,sharp corners, anchor=center, xshift=-4mm},
    			txt/.style={text width=1.5cm,anchor=center},
    			ellip/.style={ellipse,scale=1.0},
    			empty/.style={draw=none},
    			dim iso/.style={opacity=0.5, isosceles triangle, sharp corners, anchor=center, xshift=-4mm, scale=1.0},
    			dim boxes/.style={draw, rectangle, opacity=0.5,
                thick,minimum height=1cm, rounded corners,
                minimum width=1cm, black, text=black,
                text width=15mm, anchor=center, align=center},
    			]
  \matrix (mat) [matrix of nodes, nodes=boxes, column sep=0.2cm, row sep=0.5cm] 
  {
                 &   &   &           &           \\ 
    |[right iso]|{Tactile data} & |[ellip]|{Tactile preprocessing} & |[round boxes]|{PFH/SDA} &           &           \\
    |[right iso]|{Visual data} & |[ellip]|{Visual preprocessing} & |[round boxes]|{Intensity profile} & |[dim boxes]|{PC} & |[dim boxes]|{ViTa Map} \\
    |[dim iso]|{Odometry}     &   &   &     |[empty]|      &           \\
  };  
  
%% Node ordering: [row- column]
%% Tactile data
\draw [very thick, black, ->](mat-2-1)--(mat-2-2);
\draw [very thick, black, ->](mat-2-2)--(mat-2-3);
\draw [very thick, black, opacity=0.5, ->](mat-2-3)-|(mat-3-4);
%%
%%
%%%%% Visual data

\draw [very thick, black, ->](mat-3-1)--(mat-3-2);
\draw [very thick, black, ->](mat-3-2)--(mat-3-3);
\draw [very thick, black, opacity=0.5, ->](mat-3-3)--(mat-3-4);
\draw [very thick, black, opacity=0.5, ->](mat-3-4)--(mat-3-5);

%%
%%%% Odometry info connections
\draw [very thick, black, opacity=0.5, -](mat-4-1)-|(mat-4-4);
\draw [very thick, black, opacity=0.5, ->](mat-4-4)-|(mat-3-5);
\draw [very thick, black, opacity=0.5, ->](mat-4-4.center)--(mat-3-4);

%%
%%%% Obj Map and Exp Map fusion to Vita Map
%\draw [very thick, black, ->](mat-3-5)-|(mat-4-6);
%\draw [very thick, black, ->](mat-5-5)-|(mat-4-6);

%% Bounding rectangles
% draw the rectangles
\node at(-3.0cm, 1.2cm) [right,draw=black,dashed,text opacity=1,opacity=1, minimum width=5.1cm, minimum height=6.5cm, text height=-5.2cm] {Hand-crafted feature extraction};

\end{tikzpicture}}
\caption{Overview of ViTa-SLAM, blocks that are not part of the place recognition front-end are grayed out.}
\label{fig:ViTaSLAM_Structure}
\end{figure}
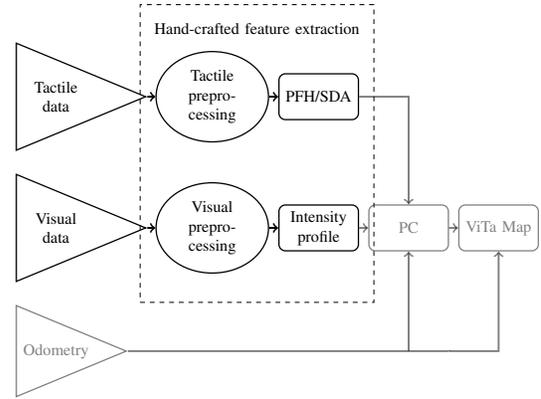

ViTa-SLAM~\cite{struckmeier2019vita} is a visuo-tactile extension to the vision-only RatSLAM~\cite{ball2013openratslam} and tactile-only WhiskerRatSLAM~\cite{salman2018whisker} methods.
ViTa-SLAM extracts visual and tactile features independently as illustrated in Fig.~\ref{fig:ViTaSLAM_Structure} showing a block diagram of ViTa-SLAM with the place-recognition front-end highlighted by the dashed square. The visual feature is an intensity profile represented as a vector $V$ and the tactile data are represented using a point feature histogram $PFH$ and a slope distribution array $SDA$~\cite{struckmeier2019vita}. 
Distance between features is defined as a weighted combination of $L_1$ differences as
\begin{equation}
\begin{split}
\epsilon_{i,j} &= \alpha~|V_i-V_j|_{L_1} + \beta~|PFH_i-PFH_j|_{L_1}\\ &+ \gamma~|SDA_i -SDA_j|_{L_1},\\
\end{split}
\label{eq:combinederror}
\end{equation}
where
\begin{equation}
\begin{split}
\alpha &= \frac{1}{\sigma_V},~\beta = \frac{1}{\sigma_{PFH}},~\gamma = \frac{1}{\sigma_{SDA}}.\\
\end{split}
\label{eq:scaling}
\end{equation}
are scaling factors to normalize the respective distances based on their standard deviations.

This hand-crafted approach for combining coincident visuo-tactile sensory information was demonstrated to be beneficial in determining place within visually aliased environments.
However, the generality was limited between different environments and experimental conditions for the following reasons:
\begin{itemize}
\item The scaling factors to normalize the error components shown in Eq.~\eqref{eq:scaling} had to be determined empirically.
\item A large number of parameters related to feature extraction and pre-processing had to be tuned.
\end{itemize}

%% Novelty
\subsection{Predictive coding for unsupervised feature extraction}
To overcome the above mentioned limitations of hand-crafted features, in this work, we propose \Method~which stands for \MethodFull. \Method~is intended to replace the place recognition front-end in ViTa-SLAM (see Fig. \ref{fig:ViTaSLAM_Structure}) with a predictive coding network which is described next.

Predicitve coding was originally developed for inferring representations of a given visual input \cite{rao1999predictive, kurkova_deep_2018}. Here, we extend it to infer unified multi-sensory representations given bimodal sensory inputs, $(\mathbf{x}^V_1, \mathbf{x}^T_1), \ldots, (\mathbf{x}^V_i, \mathbf{x}^T_i), \ldots$, where $(V)$ represents visual and $(T)$ represents tactile modalities while $i$ represents inputs.

\subsubsection{Predictive Coding Network Architecture}
Fig.~\ref{fig:high_level} shows the architecture of the \Method. The network consists of three modules, namely the visual module, tactile module and multi-sensory module. The visual module processes visual information and consists of a neural network with $N_V$ layers. Activity of the $l^{th}$ layer neurons for the $i^{th}$ input is denoted by $\mathbf{y}^{V(l)}_i$. Each layer in the network predicts the activity of the preceding layer according to
\begin{equation}
	\mathbf{\hat{y}}^{V(l-1)}_i = \phi \left( \left( \mathbf{y}^{V(l)}_i \right)^T \mathbf{W}^V_{l(l-1)} \right)^T
	\label{eq:prediction}
\end{equation}
where $\mathbf{W}^V_{l(l-1)}$ denotes the synaptic weights of the projections between the $l^{th}$ and $(l-1)^{th}$ layer in the visual module and $\phi$ is the activation function of the neurons. The lowest layer in the network predicts the visual input $(\mathbf{x}^V_i)$. Note that all layers in the network propagate information to the preceeding layer (right to left) in parallel using Eq.~\eqref{eq:prediction}. This aspect of the network is different from commonly employed feedforward networks in machine learning, like CNNs, in which information is sequentially propagated from the leftmost to rightmost layer of the network.

The tactile module consists of a similar neural network with $N_T$ layers that process tactile information. The multi-sensory module consists of a single layer which predicts the activities of neurons in the last layers of both the visual and tactile modules. The activity of neurons in this layer is denoted by $\mathbf{y}^{D}_i$ for the $i^{th}$ input and is used as features for place recognition.

\subsubsection{Learning Algorithm}
Predictive coding is used to update the synaptic weights and infer neuronal activities in the network. The $l^{th}$ layer in the visual module generates a prediction about the neuronal activities in the $(l-1)^{th}$ layer and also receives a prediction of its own neuronal activity from the $(l+1)^{th}$ layer. The goal of the learning algorithm is to infer $l^{th}$ layer neuronal activity $(\mathbf{y}^{V(l)}_i)$ for the $i^{th}$ input that generates better predictions about neuronal activity in the $(l-1)^{th}$ layer and is predictable by the $(l+1)^{th}$ layer. For this purpose, $\mathbf{y}^{V(l)}_i$ is updated by performing gradient descent on the error function
\begin{equation}
	\mathbf{e}^{V(l)}_i = \left( \mathbf{\hat{y}}^{V(l-1)}_i - \mathbf{y}^{V(l-1)}_i \right)^2 + \left( \mathbf{\hat{y}}^{V(l)}_i - \mathbf{y}^{V(l)}_i \right)^2,
\end{equation}
which results in the following update rule for $\mathbf{y}^{V(l)}_i$
\begin{equation}
	\begin{split}
		\Delta \mathbf{y}^{V(l)}_i &= \eta_y \left( \mathbf{W}^V_{l(l-1)} \left( \mathbf{y}^{V(l-1)}_i - \mathbf{\hat{y}}^{V(l-1)}_i \right) \right. \\
		&\left. + \left( \mathbf{y}^{V(l)}_i - \mathbf{\hat{y}}^{V(l)}_i \right) \right),
	\end{split}
	\label{eq:y_update}
\end{equation}
where $\eta_y$ is the learning rate for updating neuronal activities. The update rule in Eq.~\eqref{eq:y_update} is used to infer neuronal activity in all layers of the visual module for all inputs. Weights $(\mathbf{W}^V_{l(l-1)})$ between $l^{th}$ and $(l-1)^{th}$ layers in the network are updated by performing gradient descent on the error in the prediction generated by the $l^{th}$ layer neurons which results in the update rule for weights
\begin{equation}
	\Delta \mathbf{W}^V_{l(l-1)} = \eta_w \mathbf{y}^{V(l)}_i \left( \mathbf{y}^{V(l-1)}_i - \mathbf{\hat{y}}^{V(l-1)}_i \right)^T
	\label{eq:w_update}
\end{equation}
where $\eta_w$ is the learning rate for updating weights.

The learning approach for the tactile module is identical to the visual module. In case of the multi-sensory module, the representations are inferred based on prediction errors of topmost layers in both the visual and tactile modules.

\subsubsection{Feature matching}
The pairwise distance between features $\mathbf{y}_i^D$ and $\mathbf{y}_j^D$ is defined as the L2-norm as
\begin{equation}
\epsilon_{i,j} = ||(\mathbf{y}_i^D - {\mathbf{y}_j}^D)||_2 .
\label{eq:combinederror_pred}
\end{equation}

%% Empirical Evaluation
\section{Empirical Evaluation}
In this section, we describe the robot platform and the three environments used for evaluating the place recognition performance of \Method. Details of how the predictive coding network was trained and evaluated are also presented.

\begin{figure}[!tbp]
\centering
\hspace*{-0.5cm}
\begin{subfigure}[t]{0.25\textwidth}
  \centering
  \includegraphics[scale=0.06]{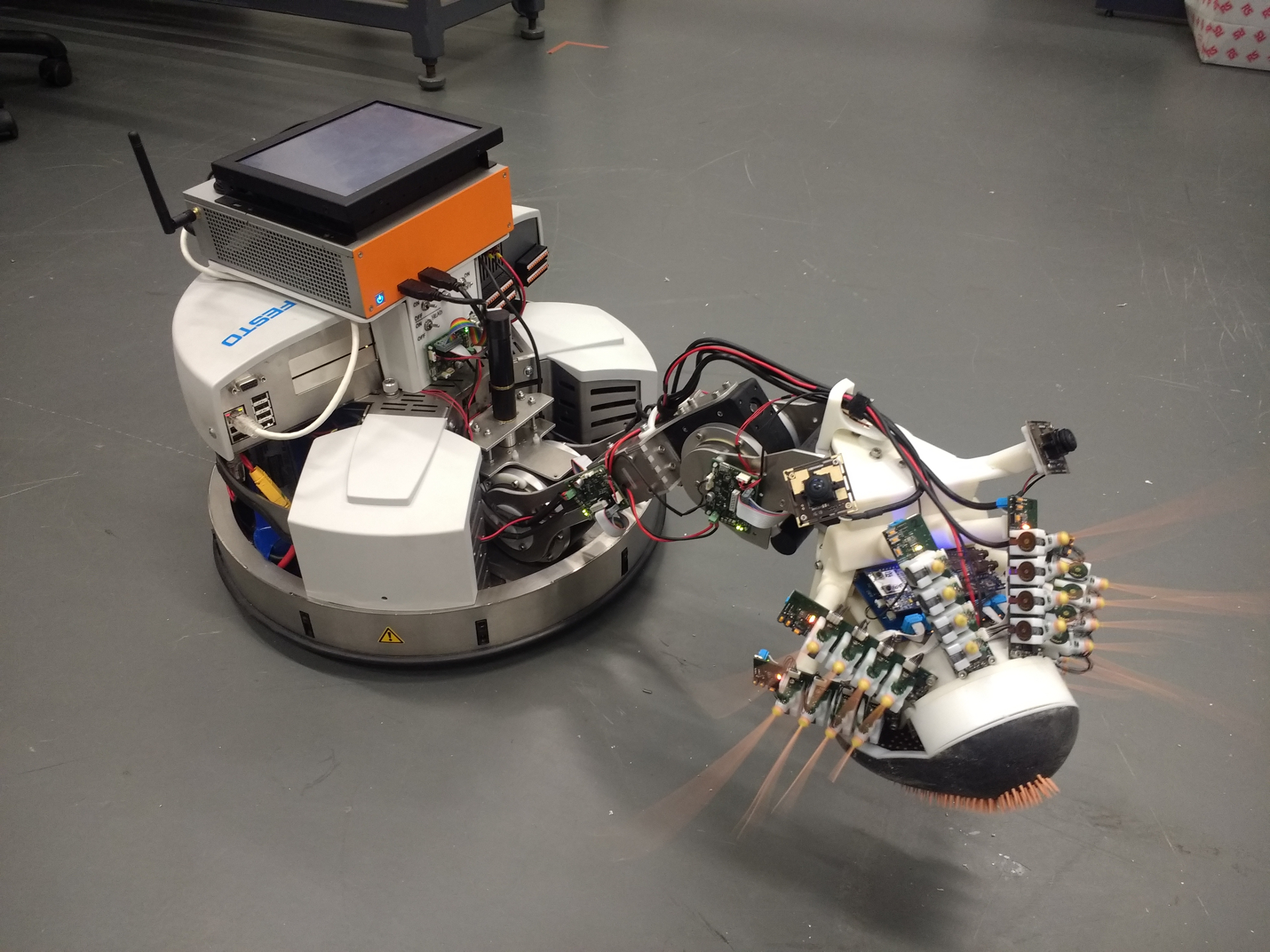}
        \caption{Physical platform.}
        \label{fig:real_whiskeye}
\end{subfigure}%
% \hspace*{-0.4cm}
\begin{subfigure}[t]{0.25\textwidth}
  \centering
  \includegraphics[scale=0.1]{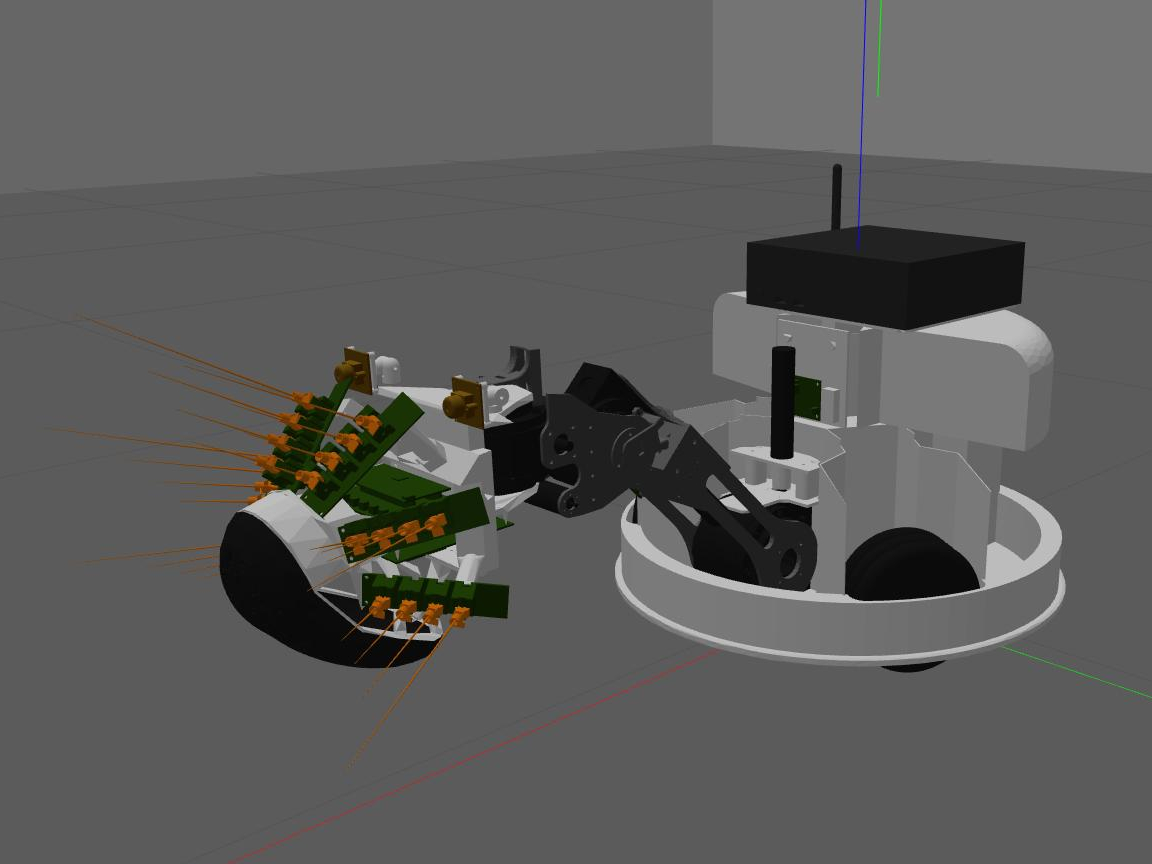}
  \caption{Simulated platform.}
  \label{fig:sim_whiskeye}
\end{subfigure}
\caption{The WhiskEye robot platform used to explore the environments shown in Fig. \ref{fig:environments} and generate the visuo-tactile data sets for training and testing of \Method.}
\label{fig:robot_platform}
\end{figure}

\begin{figure*}[!htbp]
\centering
\begin{subfigure}[t]{.32\textwidth}
  \centering
  \includegraphics[width=\textwidth]{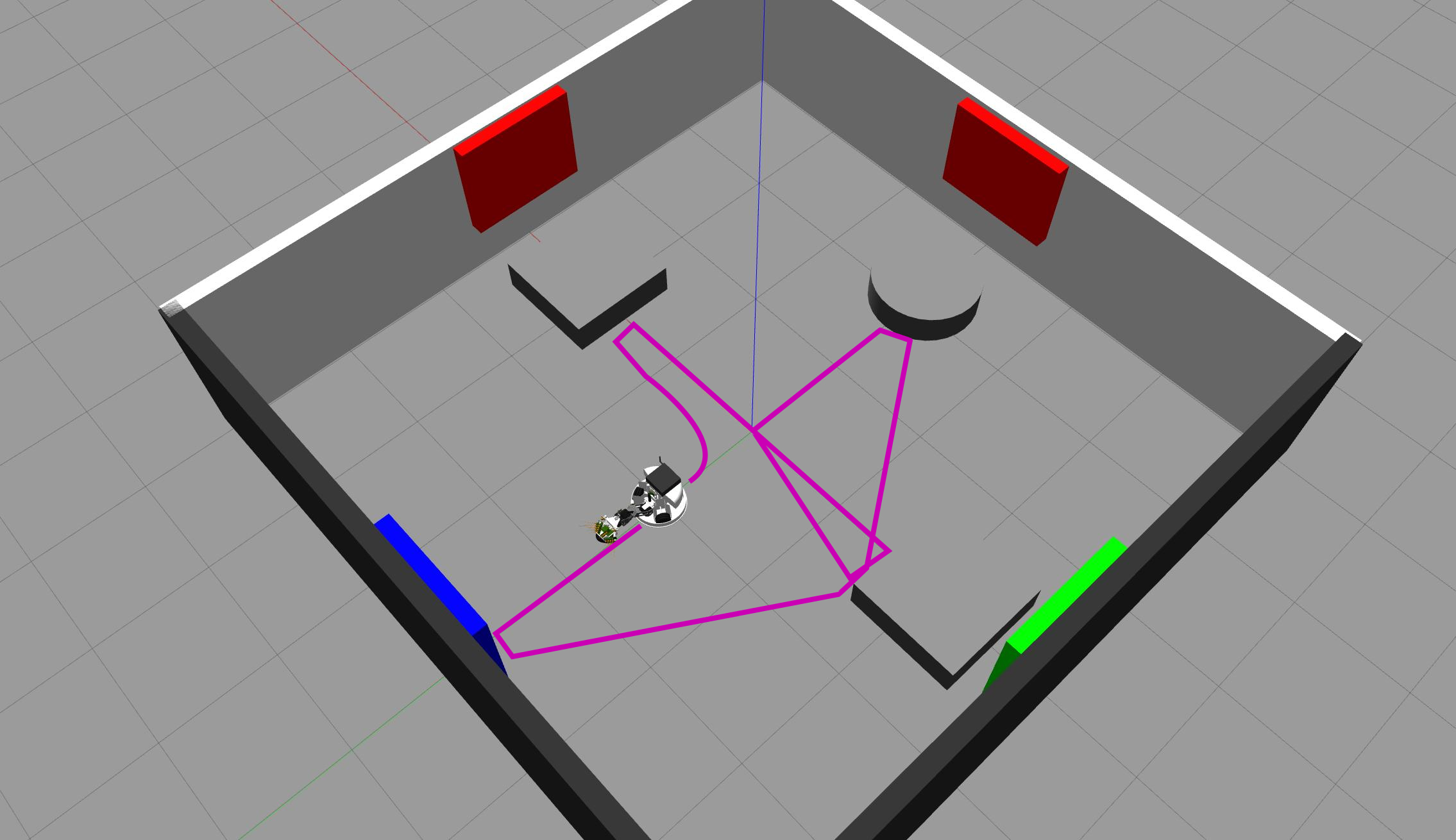}
        \caption{$E1$.}
        \label{fig:E1}
\end{subfigure}%
\hspace*{+0.01cm}
\begin{subfigure}[t]{.32\textwidth}
  \centering
  \includegraphics[width=\textwidth]{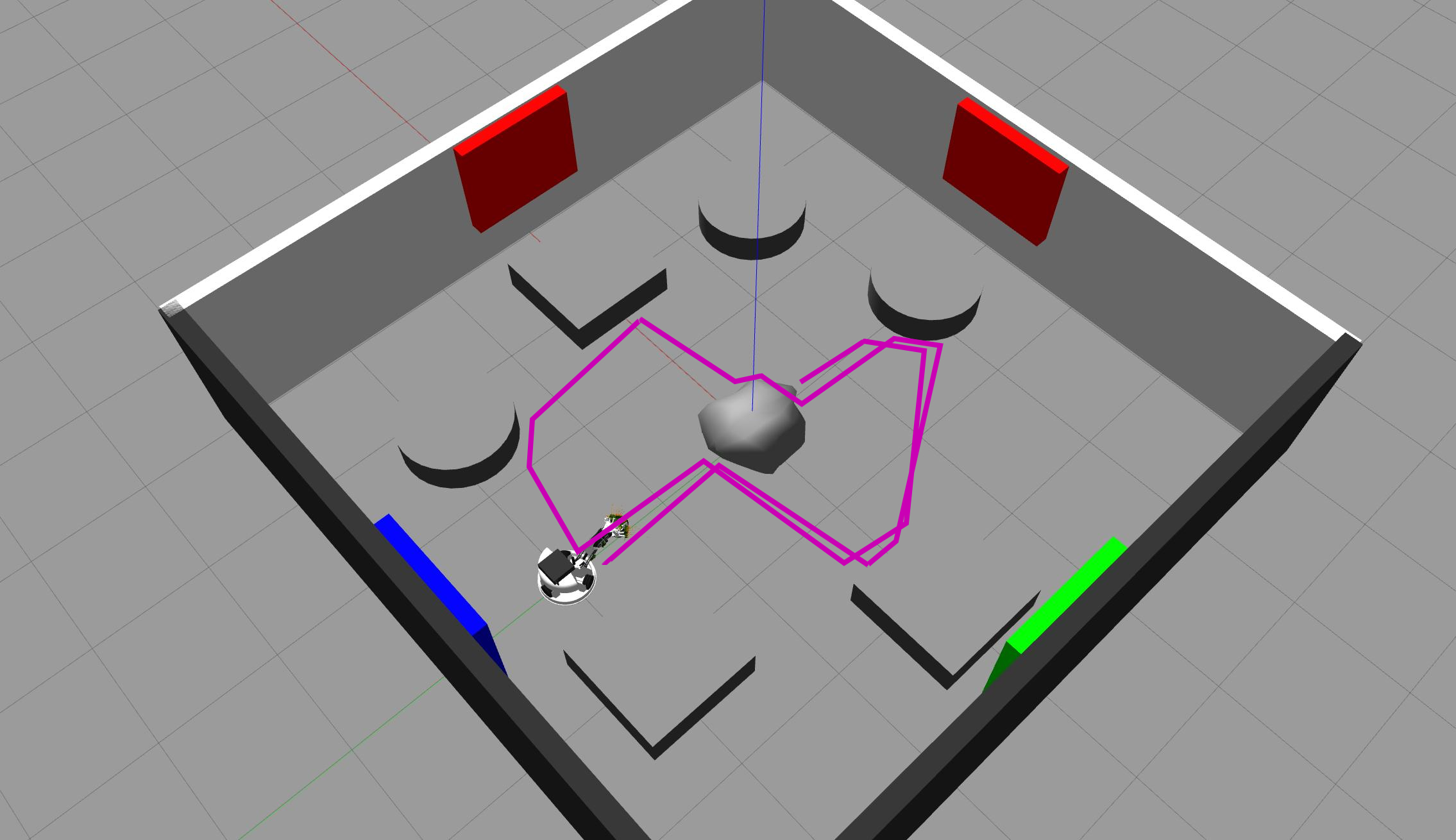}
    \caption{$E2$.}
  \label{fig:E2}
\end{subfigure}
\begin{subfigure}[t]{.32\textwidth}
    \centering
    \includegraphics[width=\textwidth]{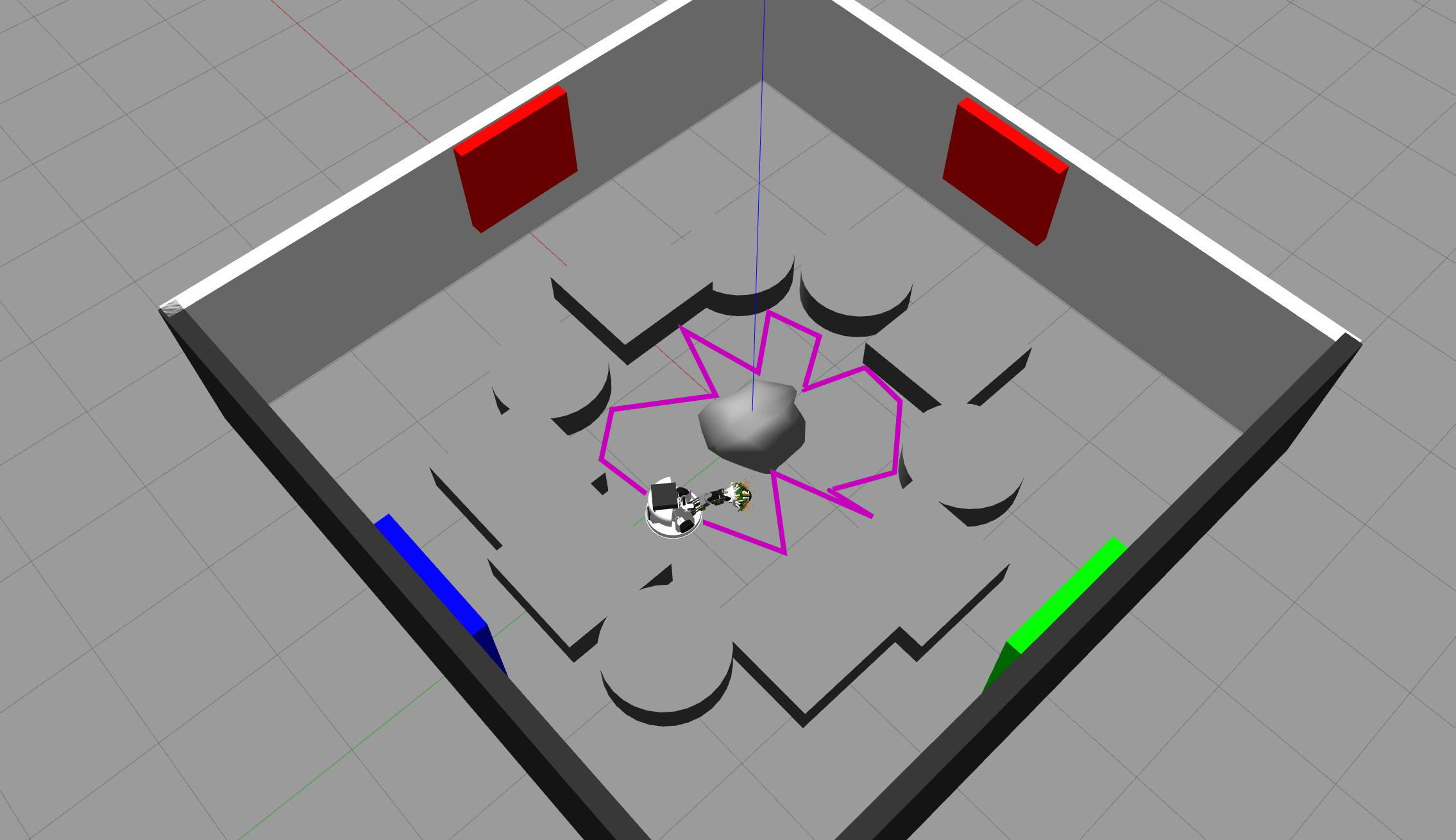}
\caption{$E3$.}
    \label{fig:E3}
\end{subfigure}
\caption{The three experiment environments with different amounts of tactile landmarks. Environment 1 has been used to gather training data for the predictive coding network. The trajectories used in the experiment are marked in magenta.}
\label{fig:environments}
\end{figure*}

\subsection{Robot Platform and Environments}
The robot platform used for this research is called the WhiskEye (Fig.~\ref{fig:real_whiskeye}). A Gazebo simulation\cite{gazebo.2004} of the WhiskEye platform (shown in Fig.~\ref{fig:sim_whiskeye}) was used to evaluate the place recognition performance similar to our previous work in \cite{struckmeier2019vita}. Mounted on the head are the visual and tactile sensors consisting of two monocular cameras with a resolution of $640 \times 480$ pixels sampled at $5$ frames per second and an array of $24$ individually actuated artificial whiskers arranged into $4$ rows of $6$.
Each whisker is instrumented with a $2$-axis hall effect sensor to detect $2$D deflections of the whisker shaft measured at its base, constituting the tactile data generated by the array\footnote{For further details about the robot platform the readers are refered to~\cite{struckmeier2019vita}.}. The whiskers are swept back and forth during exploration mimicking the \textit{whisking} behaviour observed in rats and other small mammals. The tactile data from the whiskers is extracted during every whisk cycle, at the point of maximum protraction.

The following three environments shown in Fig.~\ref{fig:environments} were chosen to vary the amount of possible tactile data that can be generated during exploration:
\begin{enumerate}
\item Environment $E1$: Identical to the environment used in \cite{struckmeier2019vita} with aliased visual and tactile landmarks.
\item Environment $E2$: A similar visual environment to $E1$, but with additional free standing tactile landmarks and a novel tactile landmark, an asymmetric rock, in the center.
\item Environment $E3$: Containing many tactile landmarks forming a continuous structure of tactile landmarks around the rock in the center.
\end{enumerate}

\subsection{Training the predictive coding network}
The \Method~was trained with data gathered as the robot explored environment $E1$. The trajectory of the robot was executed using a model of tactile attention inspired by rodent foraging behaviour \cite{mitchinson2013whisker} to generate a rich dataset. The dataset consisted of $5550$ images and whisker deflection vectors sampled in concert with visual images.
Both, tactile module and multi-sensory module had one layer with $100$ and $200$ neurons, respectively.
The visual module consisted of two layers with $1000$ and $300$ neurons, respectively.
The model was trained on NVIDIA $1080$Ti GPUs. Training for $10000$ iterations takes $5-6$ hours approximately. 

Training the \Method~involved presenting the network described in Fig. \ref{fig:high_level} with a mini-batch of $150$ concurrently recorded visuo-tactile input. For each sample in the mini-batch, the representations in each layer of the three modules in the predictive coding network were updated in parallel using Eq.~\eqref{eq:y_update}. After updating the representations, the network weights were updated using Eq.~\eqref{eq:w_update}. At the beginning of training, representations for all inputs were initialized to $0.1$ and the aim of the model was to iteratively infer representations that could decode the original sensory input. Such an initialization of representations alleviated the need for encoders in predictive coding. A single training iteration included repeating this procedure for each mini-batch. The network was trained for $10000$ iterations. The learning rates $\eta_y$ and $\eta_w$ were set to $4\times10e^{-4}$ for all layers in the network.

\subsection{Testing the predictive coding network}
During the experiments two trajectories using teleoperation through each environment were recorded to induce system noise that was crucial for the evaluation of the robustness of \Method.
To test the generalization capability of \Method, $E2$ and $E3$ were presented as novel environments during the empirical evaluations.

For testing the trained \Method, a visuo-tactile input was presented to the network and each layer in the network adapted representations in parallel using Eq.~\eqref{eq:y_update}.
The weights in the network were not adapted during testing.
This procedure was repeated for $3000$ iterations or until stimulus decoding error was lower than a user-defined threshold for all layers.

\subsection{Evaluation Metrics}
The place recognition performance was evaluated by comparing the \Method~to the place recognition front-end of ViTa-SLAM. This was done by computing the precision-recall rate as a measure for how well the methods match the templates against their spatial proximity which was taken as the ground truth.

A similar analysis  has been done in \cite{kazmi2016gist}, where Gist features were combined with a self-organizing map and used as the place recognition front-end for RatSLAM~\cite{ball2013openratslam}.
To demonstrate the similarity between vision-only scenes based on the Gist features a distance matrix was created, which contains the similarity measure for all the scenes from a trial run and it helps visualize which scenes lead to place recognitions.

These place recognitions can then be classified as: true-positive (TP), false-positive (FP) and false-negative (FN) place recognitions which are then used to compute the precision-recall rate~\cite{flach2015precision}.
Then, the precision-recall rate is computed as:

\begin{equation}
Precision = \frac{\#TP}{\#TP + \#FP};~Recall = \frac{\#TP}{\#FP + \#FN}.
\label{eq:precision_recall}
\end{equation}
The method in \cite{kazmi2016gist} focused on improving the vision-only place recognition front-end of RatSLAM and was thus able to use a dataset introduced in the work by Ball et al.~\cite{ball2013openratslam}.
This dataset did not contain ground truth pose information and thus, no automatic place recognition detection was possible.
Therefore, the dataset was manually divided in visited or unvisited scenes and the place recognition classes were determined.

Similar to \cite{kazmi2016gist}, we computed the F1-scores to compare the overall performance of each method as:
\begin{equation}
F1 $-score$ = 2\times\frac{Precision \times Recall}{Precision + Recall}.
\label{eq:f1score}
\end{equation}

\begin{figure*}[!htbp]
\centering
\begin{subfigure}[t]{.29\textwidth}
  \centering
  \includegraphics[width=\textwidth,trim={0.5cm 1cm 0.5cm 1cm},clip]{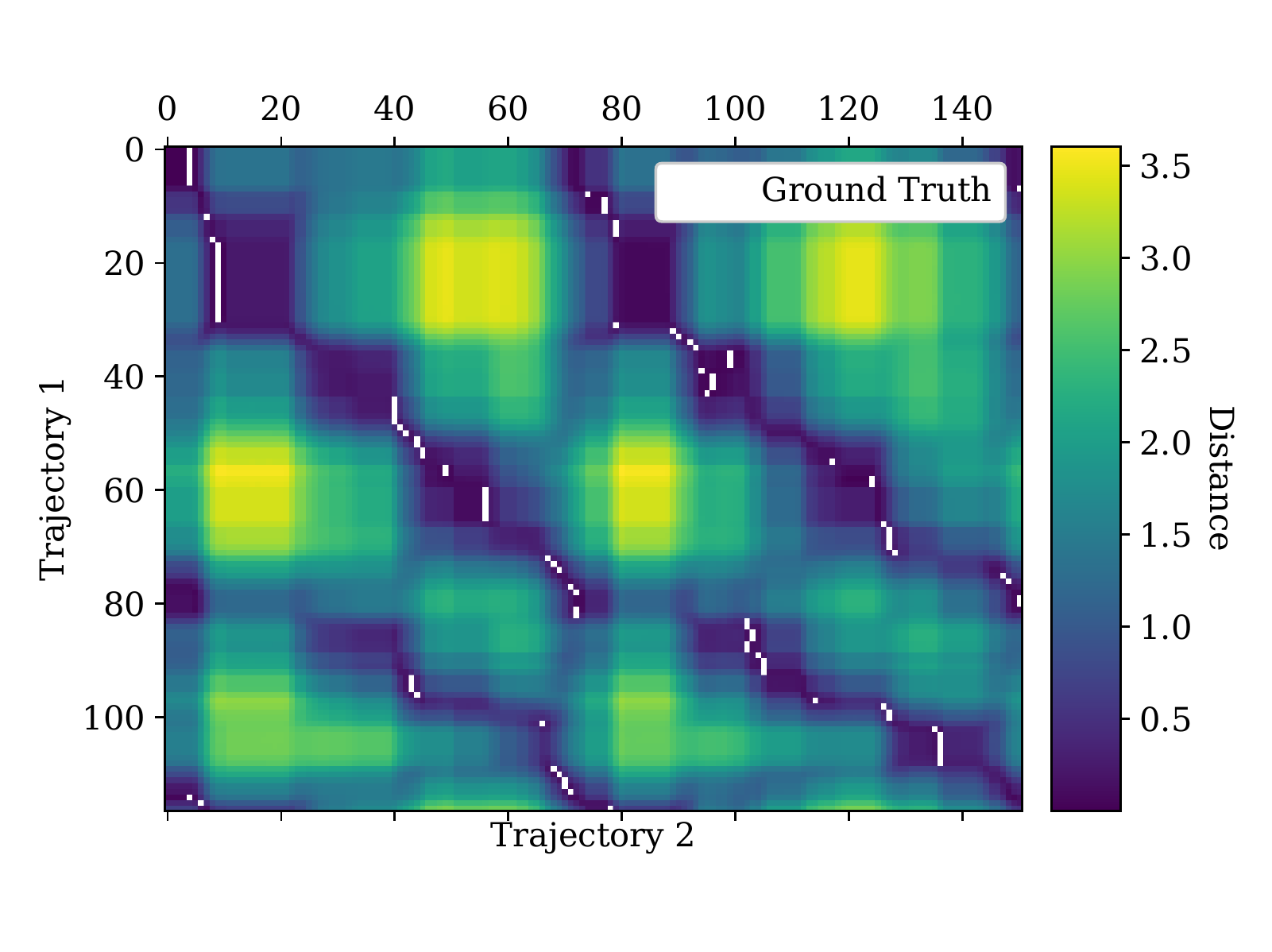}
        \caption{GTM $E1$}
        \label{fig:GTM1}
\end{subfigure}%
\begin{subfigure}[t]{.29\textwidth}
  \centering
  \includegraphics[width=\textwidth,trim={1.5cm 0cm 2.5cm 0.4cm},clip]{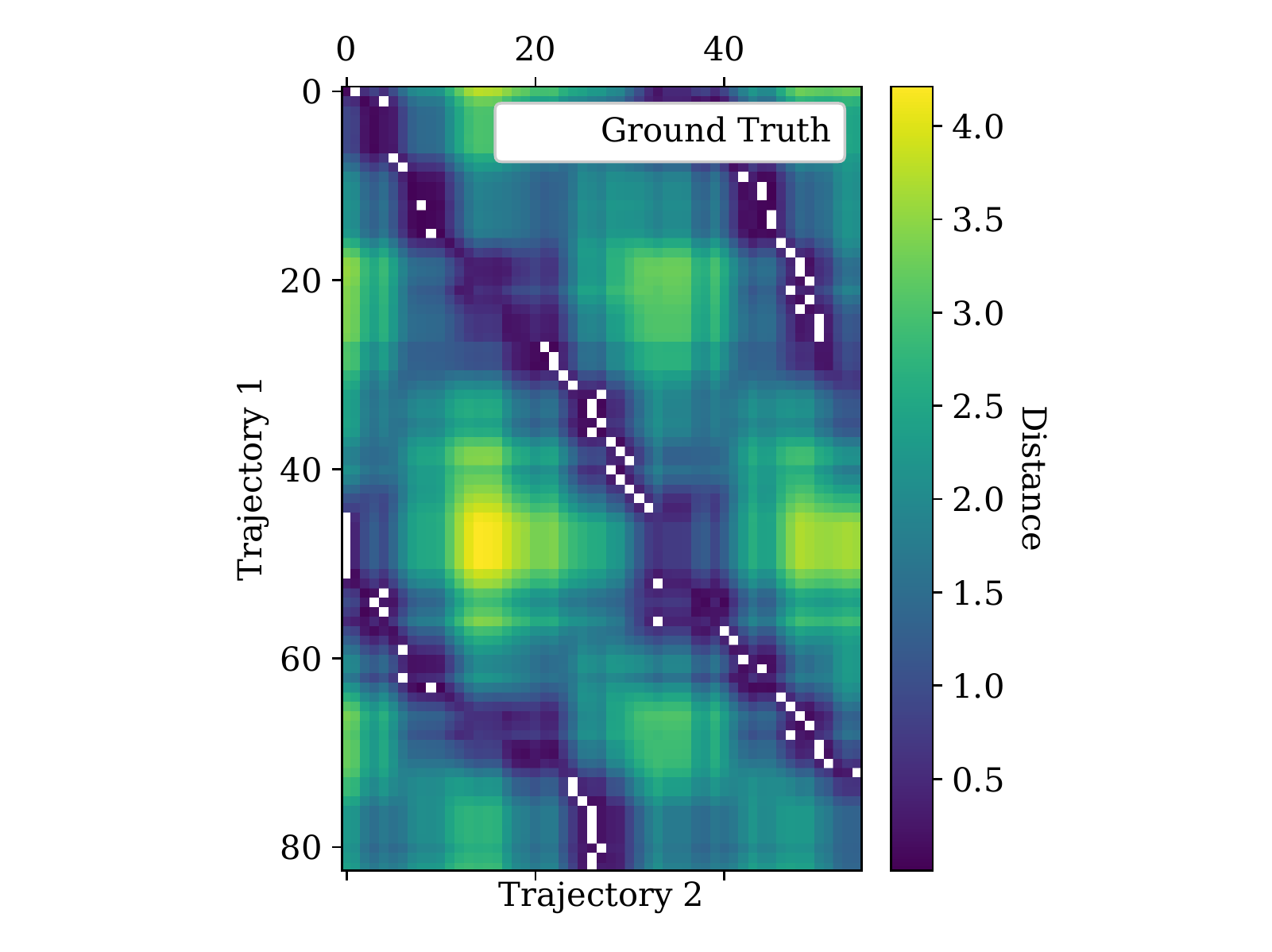}
  \caption{GTM $E2$}
  \label{fig:GTM2}
\end{subfigure}
\begin{subfigure}[t]{.29\textwidth}
    \centering
    \includegraphics[width=\textwidth,trim={2.5cm 0cm 1cm 0cm},clip]{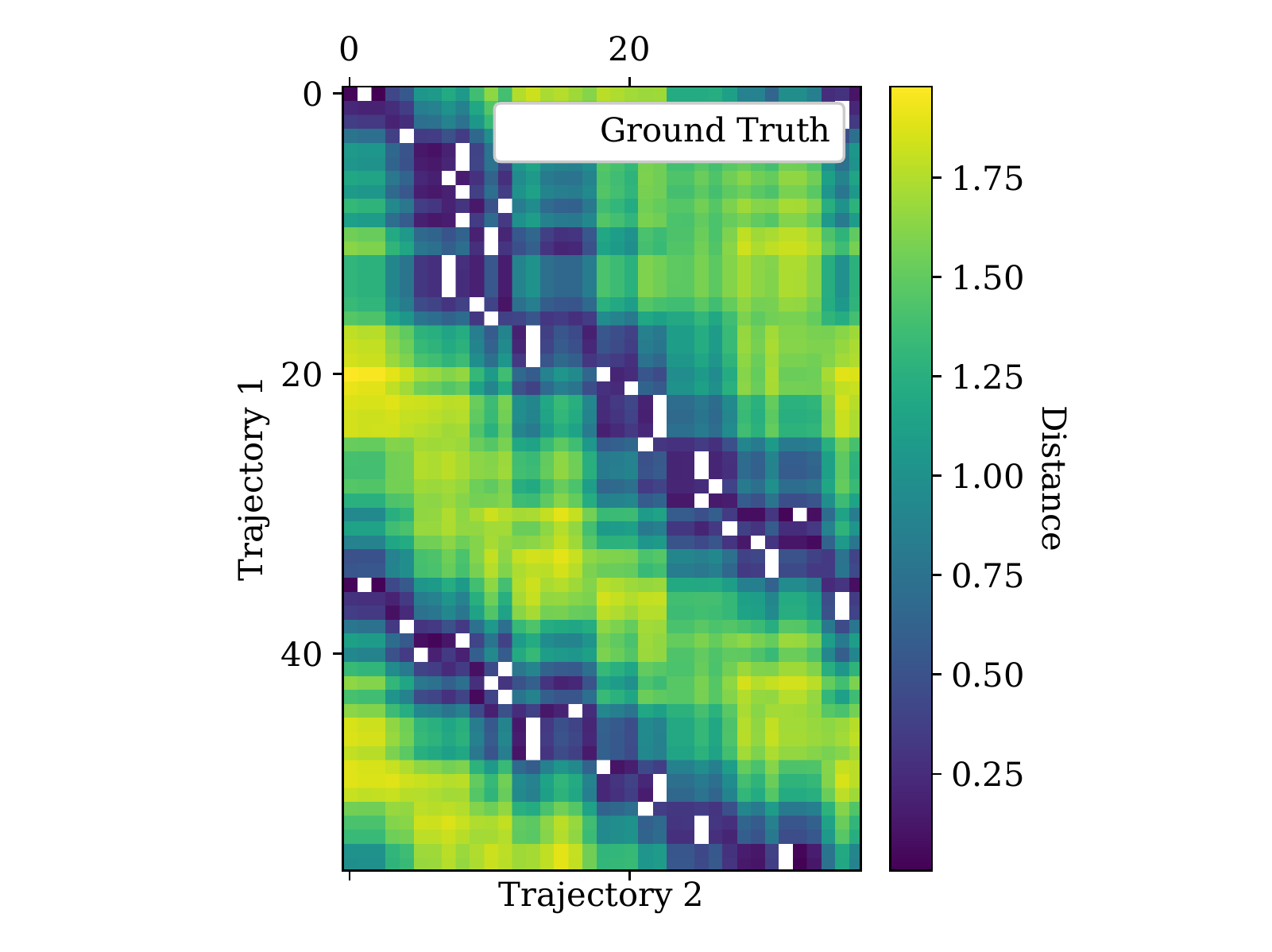}
    \caption{GTM $E3$}
    \label{fig:GTM3}
\end{subfigure}
\\
\begin{subfigure}[t]{.29\textwidth}
  \centering
  \includegraphics[width=\textwidth,trim={0.5cm 1cm 0.5cm 1cm},clip]{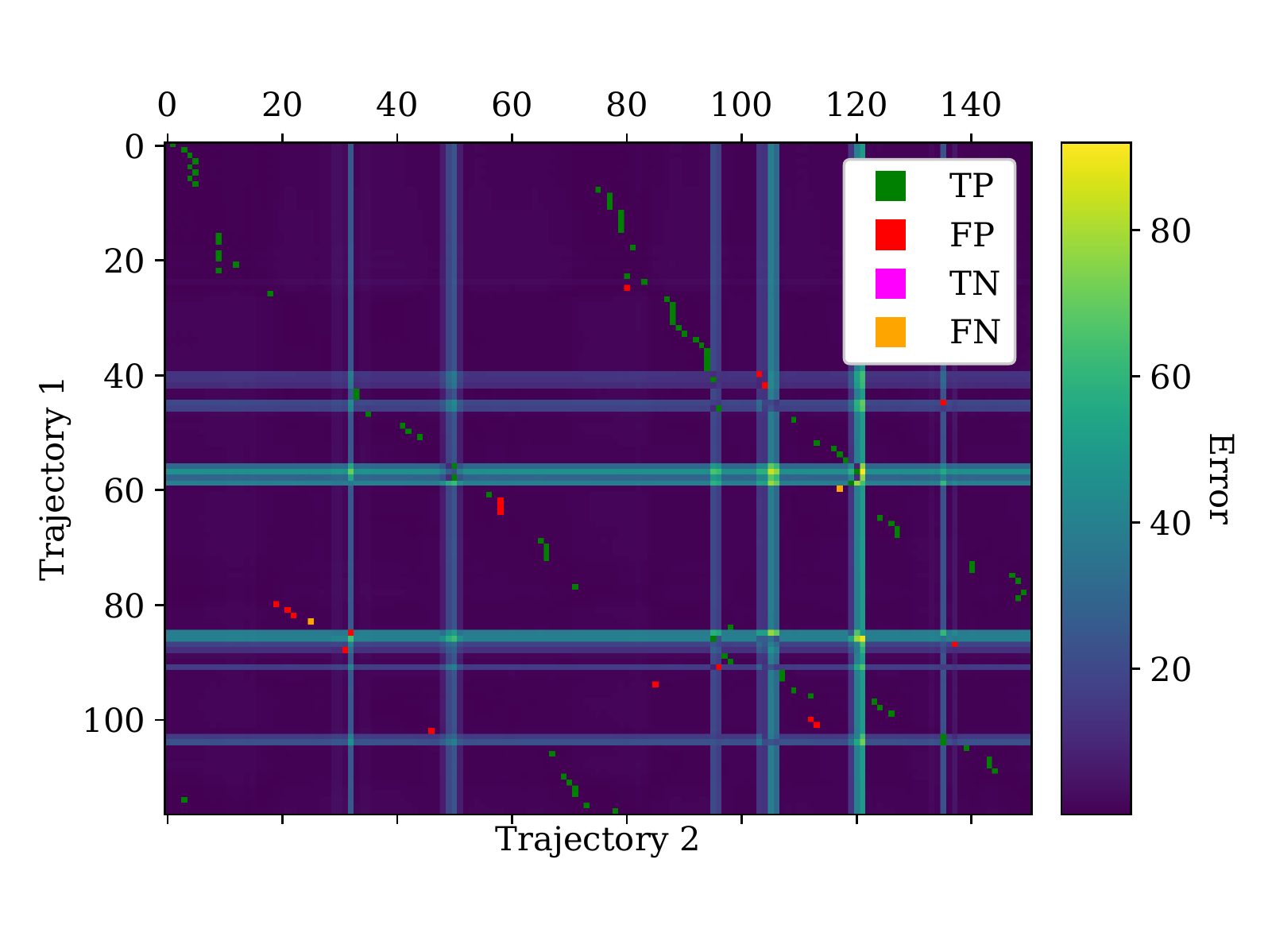}
        \caption{TME $E1 - H$.}
        \label{fig:TME1H}
\end{subfigure}%
\begin{subfigure}[t]{.29\textwidth}
  \centering
  \includegraphics[width=\textwidth,trim={1.5cm 0cm 2.5cm 0.4cm},clip]{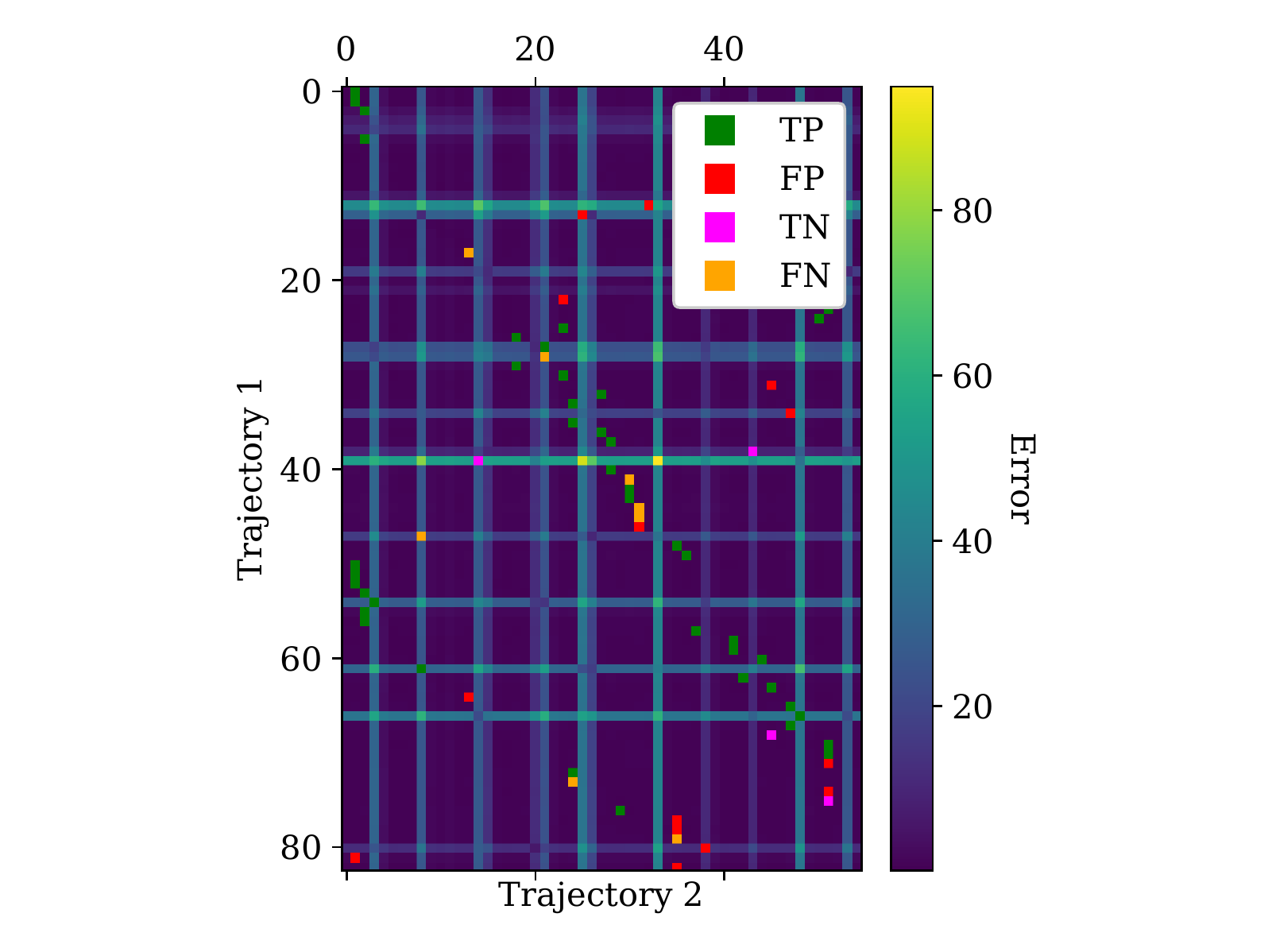}
  \caption{TME $E2 - H$.}
  \label{fig:TME2H}
\end{subfigure}
\begin{subfigure}[t]{.29\textwidth}
    \centering
    \includegraphics[width=\textwidth,trim={2.5cm 0cm 1cm 0cm},clip]{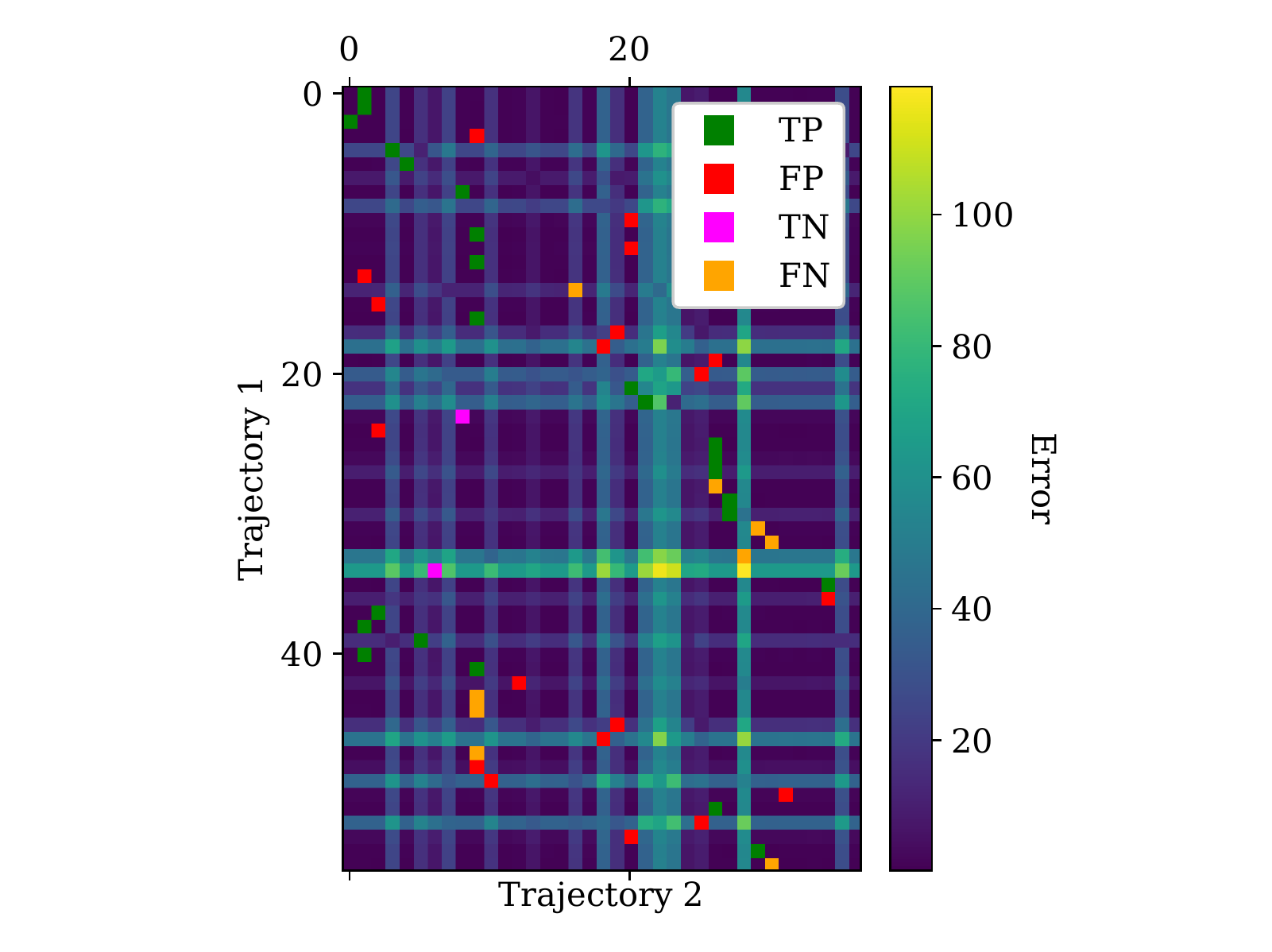}
    \caption{TME $E3 - H$.}
    \label{fig:TME3H}
\end{subfigure}
\\
\begin{subfigure}[t]{.29\textwidth}
  \centering
  \includegraphics[width=\textwidth,trim={0.5cm 1cm 0.5cm 1cm},clip]{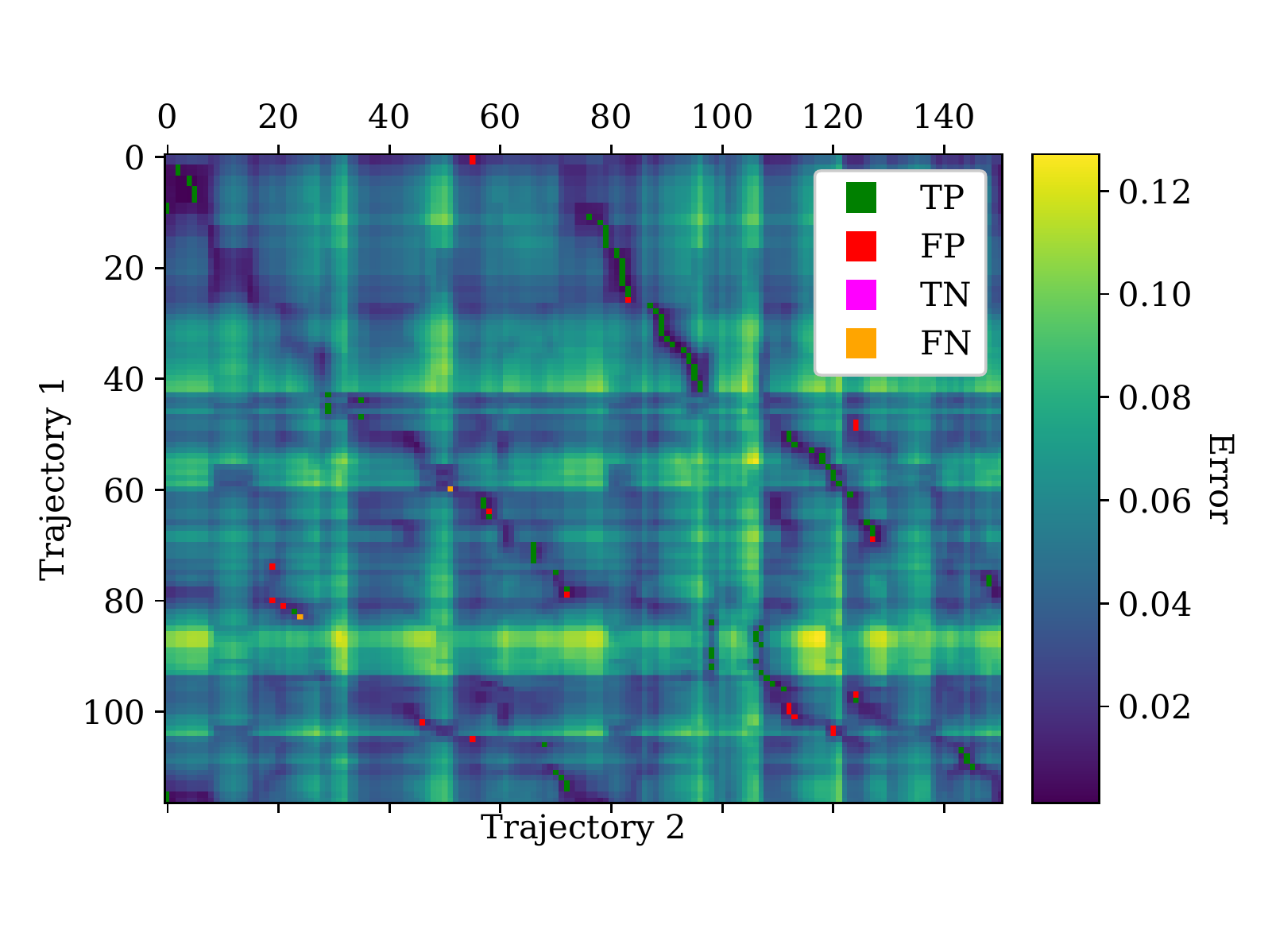}
        \caption{TME $E1 - L$.}
        \label{fig:TME1L}
\end{subfigure}%
\begin{subfigure}[t]{.29\textwidth}
  \centering
  \includegraphics[width=\textwidth,trim={1.5cm 0cm 2.3cm 0.4cm},clip]{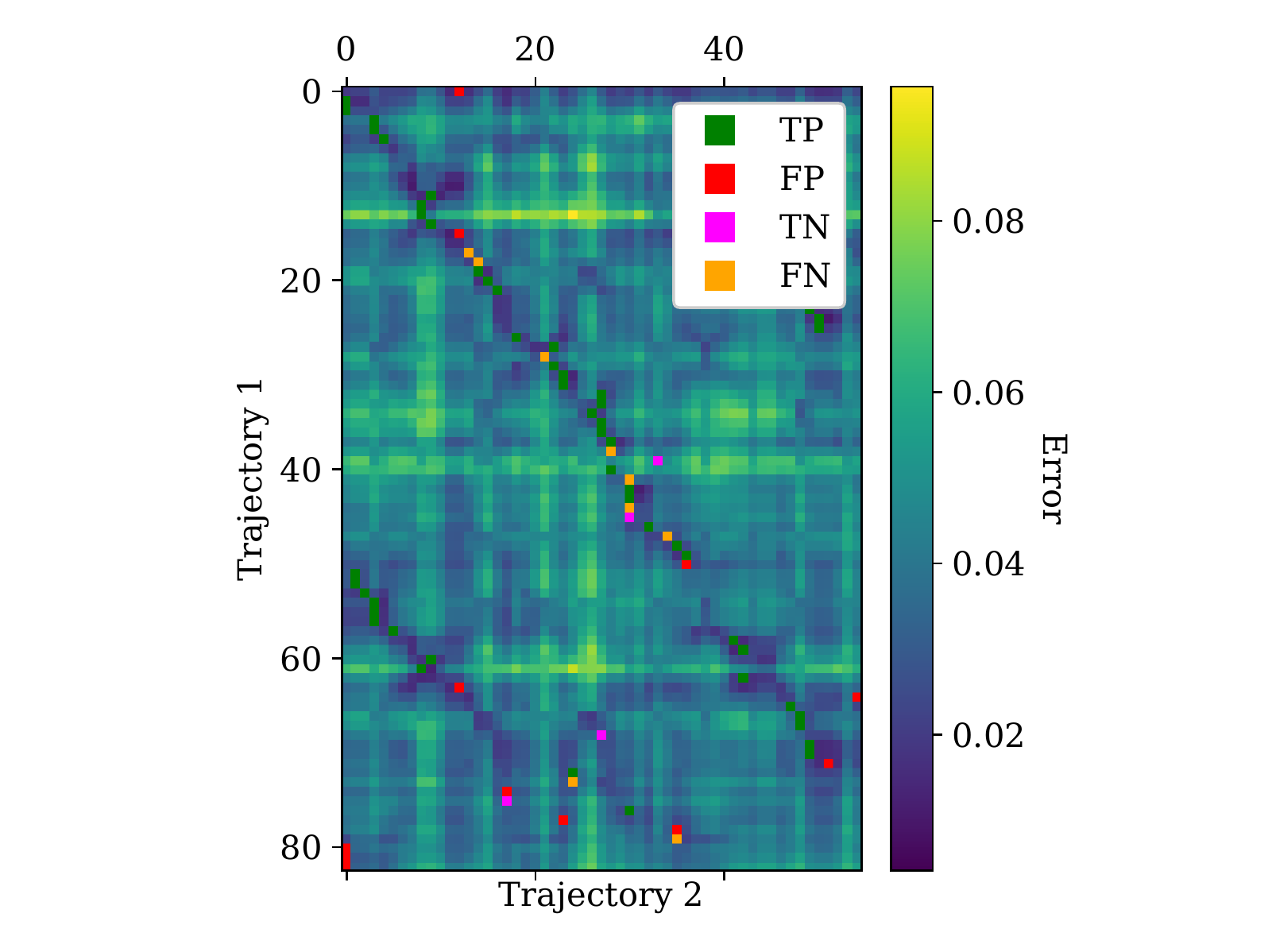}
  \caption{TME $E2 - L$.}
  \label{fig:TME2L}
\end{subfigure}
\begin{subfigure}[t]{.29\textwidth}
    \centering
    \includegraphics[width=\textwidth,trim={2.5cm 0cm 1cm 0cm},clip]{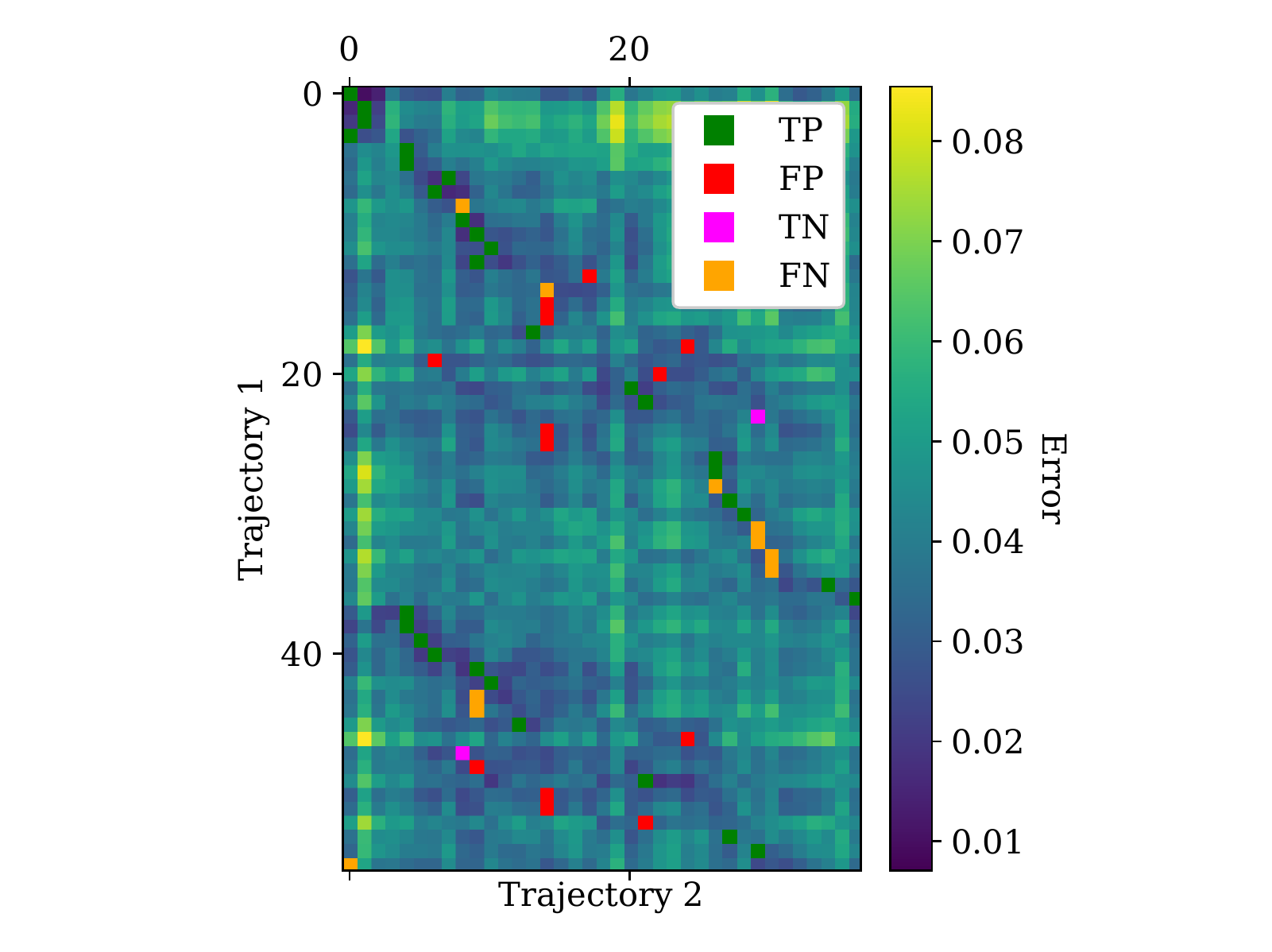}
    \caption{TME $E3 - L$.}
    \label{fig:TME3L}
\end{subfigure}
\caption{Ground truth matrices and template match error matrices for various runs using both methods: learned features (L) and hand-crafted features (H).}
\label{fig:matrices}
\end{figure*}

The differences between our method and the work in \cite{kazmi2016gist} are the availability of the multi-sensory data and the ground truth poses.
As a result, our method of detecting the place recognition types differs from the evaluation in previous work in two ways.
\textit{First}, hand-labelling of loop closures is not feasible because the tactile data is too difficult to reliably label for humans.
\textit{Second}, the poses obtained from the simulator allows us to detect ground truth place recognition events automatically.
Thus, to obtain the TP, FP and FN place recognitions, the \textit{best matches}\footnote{The best match of templates are the pair yielding the smallest template difference as explained in Eqs.~\eqref{eq:combinederror} and \eqref{eq:combinederror_pred}.} for each pair of templates are compared to the ground truth pose difference between the two templates.
If two similar templates were also recorded in similar physical poses (within a threshold $\tau$), a TP place recognition occurred.
The similarity, as computed with Eq. \eqref{eq:combinederror_pred}, between all templates is visualized using the \textit{template match error matrix} (TME), which is similar to the distance matrix introduced in \cite{kazmi2016gist}.
The ground truth pose difference between all templates are visualized in the \textit{ground truth matrix} (GTM).
Finally, with the TP, FP and FN place recognitions, the precision-recall rate and the F1 score are computed using Eq. \eqref{eq:precision_recall} and Eq. \eqref{eq:f1score}.

The different method of determining place recognitions based on the ground truth distance comes with another challenge.
When computing the template match error matrix, the smallest template match error will always be 0 between a template and itself.
A re-visit of the same template will have a small template match error, but it will always be greater than 0 because of noise in the trajectories and sensory data.
Thus when applying our method of using ground truth pose information with one trajectory it is not possible to get nonzero template differences and pose differences.
Therefore, in our evaluation, we record two similar trajectories with random noise and us them to generate the TME and the ground truth matrix.

%% Results
\section{Results}
In this section, we discuss the performance of learned features $L$ against hand-crafted features $H$ for place recognition in the three experimental environments $E1$ - $E3$.
The naming convention for an experiment run is as follows: place recognition using method $M$ in environment $E^*$ is termed $E^*-M$ where $M \in [L, H]$.

An intuition for the comparison of both methods can be obtained by observing at the template match error matrices and ground truth matrices as shown in Fig. \ref{fig:matrices}.
Figs. \ref{fig:GTM1} - \ref{fig:GTM3} show the ground truth matrices for each of the three environments. The white dots indicate the minimum ground truth pose difference for each template pair.
The figure displays the template match error matrices for each run, the green and magenta dots represent the true positive and true negative matches, the red and orange dots represent the false positive and false negative matches respectively.
We can see that the TMEs generated using hand-crafted features displays high spikes in template match error for templates with tactile sensory data.
With learned features the errors are more evenly distributed, indicating that learned features are able to better combine visual and tactile data.
These smooth gradients in the learned template match error matrix indicate that close spatial proximity does not lead to large variations in template match error, which can make it more difficult to find a good threshold that indicates a place recognition when performing memory recall.

Considering the error ranges, in Fig. \ref{fig:TME3H} the maximum template match errors are larger compared to Fig. \ref{fig:TME1H} and Fig. \ref{fig:TME2H}.
Further analysis of the recorded templates and hand-crafted template match errors in Fig. \ref{fig:TME3H} shows that the largest spike in template match error occur when tactile data is present.
This caused by novel visuo-tactile stimuli not present in $E1$, such as the complex wall shape constructed from overlapping tactile landmarks, which would require new tuning of the scaling factors $\alpha, \beta$ and $\gamma$.
Compared to that, the learned features display more similar ranges of errors across the different environments.
We therefore conclude that learned features can generalize better, especially in situations where multi-modal sensory input is present.
The reason is that higher dimensional visual data provides a larger variety of data to determine the scaling factor $\gamma$ such that it is better able to generalize
Introducing new tactile landmarks on the other hand can lead to very different tactile signatures caused by the low dimensionality of the whisker tactile data.
Thus, a crucial advantage of the learned features is that the tactile data is better accommodated while performing inference.

\begin{figure}[t]
\centering
\includegraphics[width=\columnwidth, ,trim={0cm 0.5cm 0cm 0cm},clip]{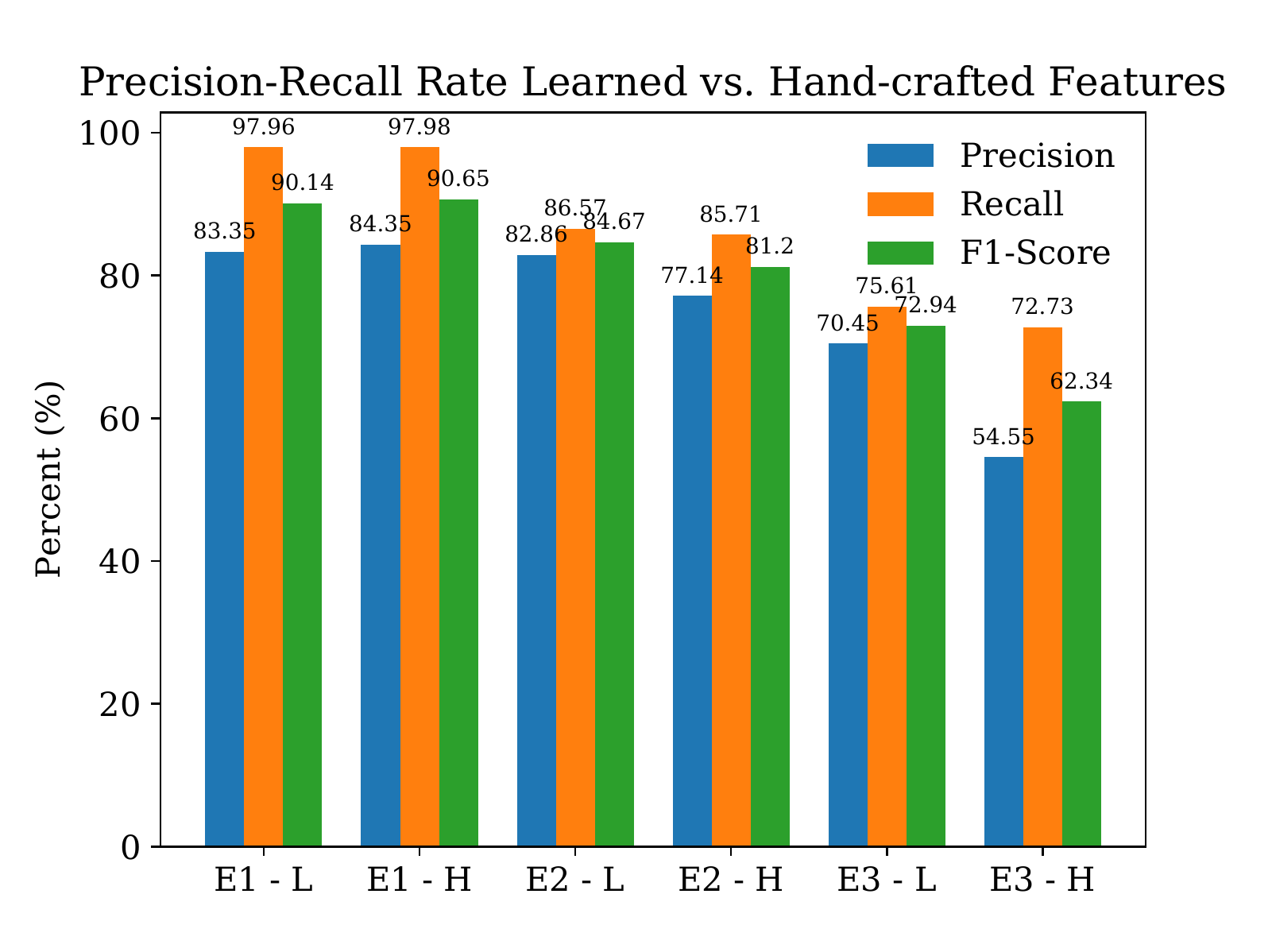}
\caption{Precision-recall rate and F1-score of both methods, learned features ($L$) and hand-crafted features ($H$) for each environment $E1$ to $E3$. In $E1$ the training data for $L$ was generated and the parameters for $H$ have been tuned. The performances in this environment are similar. In $E2$ our method reaches higher precision. In $E3$ our method clearly outperformed the hand-crafted features with a F1-score of $72.94\%$ to $62.34\%$.}
\label{fig:scores}
\end{figure}
To quantify the place recognition performance we will discuss the precision-recall rates of the learned features compared to the hand-crafted features as shown in Fig. \ref{fig:scores}.
The performance between the two methods in $E1$ was very similar and the best compared to the other environments.
This is not a surprising result given that $E1$ has been used to gather training data for method $L$ and to tune the parameters for method $H$.
In environment $E2$ the general structure of the environment was similar to $E1$ but a new tactile landmark was introduced and the amount of templates with tactile data was higher.
Performance differences are becoming apparent: while the recall performance deteriorated for both methods, method $L$ is capable of maintaining a higher precision rate.
These differences are further apparent in $E3$.
The amount of tactile data was the highest across all environments and tactile landmarks overlapped and created new tactile stimuli.
Furthermore, overlapping tactile landmarks are also harder to distinguish visually than free standing tactile landmarks, making visual data more ambiguous.
In this run, \Method~performed $10.6\%$ better than hand-crafted features, confirming the findings from analyzing the TMEs: learned features are better able to accommodate the tactile data when inferring representations.

%% Conclusion and Future works
\section{Conclusion}
In this work, we presented \Method~for biologically plausible extraction of visuo-tactile latent representations.
\Method~extends existing predictive coding approaches to multi-sensory inference.
We demonstrated that the extracted features can be used for robust place recognition.
Experiments indicated that the proposed features are superior to existing hand-crafted alternatives in novel environments in the place recognition domain.
The experimental evaluation was performed only in simulation, which may limit the generality of the findings.
However, the fact that in the proposed method the representations are learned agrees with many other current findings that learned representations are superior to hand-crafted ones.

The proposed predictive coding based approach for multi-modal feature extraction is not limited to visuo-tactile processing.
This opens interesting avenues for future research. 
Moreover, some earlier works have also shown improved performance when using a combination of hand-crafted and machine learned features \cite{nanni2017handcrafted,kan2019supervised}.
Such an approach might yield interesting insights for place recognition.
other sensory modailties

Although results with the current simplistic \Method~have shown promising improvements, further enhancement to the architecture of the predictive coding, could further improve the quality of the feature extraction.
For example convolutional layers in the visual module and recurrent structures for continuous-time inference.

%%% Bibliography
\bibliographystyle{ieeetr}
\bibliography{predictive_vita_slam}
\end{document}